\documentclass{article}
\usepackage{arxiv}

\usepackage[utf8]{inputenc} 
\usepackage[T1]{fontenc}    
\usepackage{hyperref}       
\usepackage{url}            
\usepackage{booktabs}       
\usepackage{amsfonts}       
\usepackage{nicefrac}       
\usepackage{microtype}      
\usepackage{lipsum}		
\usepackage{graphicx}
\usepackage{natbib}
\usepackage{doi}
\usepackage{subcaption}
\usepackage{adjustbox}
\usepackage{mhchem}
\usepackage{amssymb}
\usepackage{latexsym}
\usepackage{mathrsfs}
\usepackage{framed,multirow}

\title{Autonomous Adaptive Solver Selection for Chemistry Integration via Reinforcement Learning}

\author{ \href{https://orcid.org/0000-0002-6438-1389}{\includegraphics[scale=0.06]{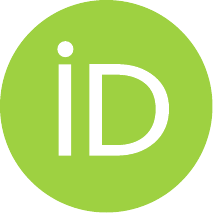}\hspace{1mm}Eloghosa Ikponmwoba} \\
	Department of Mechanical and Industrial Engineering \\
    Louisiana State University \\
    Baton Rouge, LA 70803 \\
	\texttt{eikpon1@lsu.edu} \\
	\And
	\href{https://orcid.org/0000-0001-5531-2245}{\includegraphics[scale=0.06]{orcid.pdf}\hspace{1mm}Opeoluwa Owoyele} \\
	Department of Mechanical and Industrial Engineering \\
    Louisiana State University \\
    Baton Rouge, LA 70803 \\
	\texttt{oowoyele.edu} \\
}

\renewcommand{\shorttitle}

\hypersetup{
}

\begin{document}
\maketitle

\begin{abstract}
   The computational cost of stiff chemical kinetics remains a dominant bottleneck in reacting-flow simulation, yet hybrid integration strategies are typically driven by hand-tuned heuristics or supervised predictors that make myopic decisions from instantaneous local state. We introduce a constrained reinforcement learning (RL) framework that autonomously selects between an implicit BDF integrator (CVODE) and a quasi-steady-state (QSS) solver during chemistry integration. Solver selection is cast as a Markov decision process. The agent learns trajectory-aware policies that account for how present solver choices influence downstream error accumulation, while minimizing computational cost under a user-prescribed accuracy tolerance enforced through a Lagrangian reward with online multiplier adaptation.
    Across sampled 0D homogeneous reactor conditions, the RL-adaptive policy achieves a mean speedup of approximately $3\times$, with speedups ranging from $1.11\times$ to $10.58\times$, while maintaining accurate ignition delays and species profiles for a 106-species \textit{n}-dodecane mechanism and adding approximately $1\%$ inference overhead. Without retraining, the 0D-trained policy transfers to 1D counterflow diffusion flames over strain rates $10$--$2000~\mathrm{s}^{-1}$, delivering consistent $\approx 2.2\times$ speedup relative to CVODE while preserving near-reference temperature accuracy and selecting CVODE at only $12$--$15\%$ of space-time points. Overall, the results demonstrate the potential of the proposed reinforcement learning framework to learn problem-specific integration strategies while respecting accuracy constraints, thereby opening a pathway toward adaptive, self-optimizing workflows for multiphysics systems with spatially heterogeneous stiffness.
\end{abstract}

\keywords{reinforcement learning \and combustion modeling \and chemical kinetics \and adaptive solvers \and scientific machine learning}


\section{Introduction}
\label{sec1}

High-fidelity combustion simulation has become indispensable for developing modern energy systems, including internal combustion engines, gas turbine combustors, and emerging alternative fuel technologies. Accurate prediction of ignition delay, flame structure, pollutant formation, and fuel oxidation pathways requires coupling fluid dynamics, transport processes, and multi-step chemical kinetic mechanisms. However, the chemical kinetics-related modules within reacting flow solvers typically consume 70--90\% of the total computational cost \cite{bracconi2017situ,lu2009toward,liang2009use}, creating a critical computational bottleneck. This computational burden arises from the extreme stiffness of the governing ordinary differential equation (ODE) systems, where fast radical chain-branching and recombination reactions evolve on sub-nanosecond timescales while slower fuel oxidation and thermal relaxation processes occur on millisecond scales. 
Formally, finite-difference analysis of the constant-pressure 0D chemical source-term Jacobian reveals eigenvalues whose real parts span many orders of magnitude, reflecting extreme spectral separation across fast and slow chemical timescales. This pronounced stiffness presents severe stability and performance challenges for numerical integrators and motivates the use of adaptive implicit methods~\cite{hindmarsh2005sundials}.

To address stiffness, robust implicit integrators such as the backward differentiation formula (BDF) methods implemented in CVODE (part of the SUNDIALS suite \cite{hindmarsh2005sundials}) have become the workhorse for combustion kernel solvers. The implicit time-stepping strategy enables large stable time steps even in highly stiff regimes but incurs substantial computational overhead from Jacobian evaluations, Newton-Krylov iterations, and expensive linear solver phases, particularly problematic when detailed mechanisms with hundreds to thousands of species are employed. Conversely, explicit methods or quasi-steady-state (QSS) approximations exploit timescale separation by assuming fast intermediates reach local equilibrium, thereby reducing integration cost \cite{mott2000quasi}. However, these methods become unreliable or divergent in strongly stiff or gradient-rich flame zones where the quasi-steady-state assumption breaks down \cite{flach2006use}. This fundamental trade-off (robustness versus efficiency) motivates the search for adaptive strategies that can dynamically select the appropriate solver based on local conditions.

Recognizing this challenge, researchers have developed various hybrid approaches to mitigate chemistry integration costs. Numerous works have used stiffness metrics derived from the source term \cite{shi2012accelerating} or the eigenvalues of the system Jacobian \cite{curtis1983jacobian, lu2010three} to switch between explicit and implicit solvers. Early work by Young and Boris introduced CHEMEQ \cite{young1980chemeq}, which switches between stiff and non-stiff methods based on time-scale analysis. Liang et al. \cite{liang2007development, liang2009use} proposed dynamic adaptive chemistry (DAC) that partitions the computational domain into regions requiring detailed versus reduced kinetics, achieving significant speedups in engine simulations. 

Despite these advances, most existing approaches rely on problem-specific heuristics, such as temperature thresholds, species concentration limits, or geometric domain partitioning. This usually requires manual calibration for each application. Such fixed rules lack adaptability to spatiotemporal stiffness variations that emerge naturally in reacting flows, particularly in complex configurations with moving flame fronts, extinction-reignition events, or adaptive mesh refinement. Furthermore, as combustion research increasingly focuses on alternative fuels (biodiesel, hydrogen blends, ammonia) with different kinetic characteristics, the need for solver strategies that can automatically adapt without expert re-tuning becomes paramount. A recent review by Niemeyer and Sung \cite{niemeyer2014accelerating} highlighted this gap, noting that while substantial progress has been made in mechanism reduction and tabulation, the fundamental question of when and where to apply which numerical method remains largely heuristic-driven.

In parallel, machine learning and artificial intelligence methods have begun transforming computational physics and engineering. Neural network surrogates for thermochemical property prediction \cite{blasco1998modelling, blasco2000self}, convolutional neural networks for turbulence modeling \cite{fukami2019super, kim2020prediction}, and physics-informed neural networks (PINNs) for PDE solutions \cite{nguyen2022efficient} demonstrate the potential of data-driven approaches. In combustion specifically, learned intelligent tabulation schemes \cite{haghshenas2021acceleration} replace traditional in-situ adaptive tabulation (ISAT) \cite{pope1997computationally} with neural networks, achieving faster retrieval with comparable accuracy. Owoyele and Pal \cite{owoyele2022chemnode} introduced ChemNode, a machine learning framework to predict chemical source terms during integration. These efforts demonstrate that data-driven methods can successfully accelerate chemistry computations while maintaining acceptable accuracy.

More directly relevant to solver selection, Lapointe et al. \cite{lapointe2020data} presented a data-driven approach using neural networks to predict ODE solver CPU times and errors for given thermochemical states, enabling optimal solver selection on a cell-by-cell, timestep-by-timestep basis in operator-splitting CFD simulations. However, their method requires separate neural networks for each solver to predict cost and error, relies on supervised labels from exhaustive solver sweeps during training, and uses a greedy selection strategy that does not account for long-term consequences of solver choices. Furthermore, the approach does not provide mechanisms for the policy to learn trade-offs between cost and accuracy beyond what is encoded in the pre-labeled training data. This approach relied on a supervised learning algorithm trained on partially-stirred reactor and flame data.

Reinforcement learning (RL), a branch of machine learning concerned with learning optimal policies through sequential interaction within an environment offers a natural extension of this paradigm by formulating solver selection as a sequential decision problem where an agent learns optimal policies through interaction with the environment rather than supervised labels. By casting the problem as a Markov decision process (MDP), an RL agent can discover strategies that account for temporal dependencies, for instance, recognizing that using an expensive solver now may enable cheaper integration later, or that accumulated errors over multiple steps must be managed. Recent applications of RL in scientific computing include adaptive mesh refinement \cite{yang2023reinforcement, freymuth2023swarm, foucart2023deep}, multigrid method selection \cite{luz2020learning, huergo2024reinforcement}, and turbulence model switching \cite{novati2021automating}, demonstrating feasibility for physics-based simulation tasks. Proximal Policy Optimization (PPO) \cite{schulman2017proximal}, a state-of-the-art policy gradient method, has proven particularly effective due to its stable gradient updates via clipped objectives and sample efficiency through generalized advantage estimation.

Our contributions in this paper are as follows: first, we present a RL approach for automatic solver selection in reacting flow applications. We demonstrate that reinforcement learning can discover adaptive solver selection strategies that achieve significant speedups in reacting flow applications compared to CVODE-only baselines, while maintaining good accuracy. Secondly, we demonstrate some degree of generalization: policies trained exclusively on 0D trajectories generalize to 1D counterflow flames across strain rates spanning two orders of magnitude (10--2000~s$^{-1}$) and adapt naturally to dynamic mesh refinement. Thirdly, we include a discussion suggesting that the learned strategy is physically consistent.

The remainder of this paper is organized as follows. Section~\ref{sec:method} describes the problem formulation, including the chemistry integration solvers (CVODE and QSS), MDP design, Lagrangian constraint reward function, and PPO training procedure. Section~\ref{sec:results} presents results for 0D homogeneous reactors and 1D counterflow flames, including performance metrics, generalization tests, policy interpretability analysis, and comparisons to fixed-threshold heuristics. Section~\ref{sec:conclusion} concludes with key findings and broader impacts for scientific computing applications involving spatially heterogeneous degrees of stiffness,  current limitations and future work.

\section{Methodology}
\label{sec:method}

\begin{figure*}[htbp]
    \centering
    \includegraphics[width=0.8\textwidth]{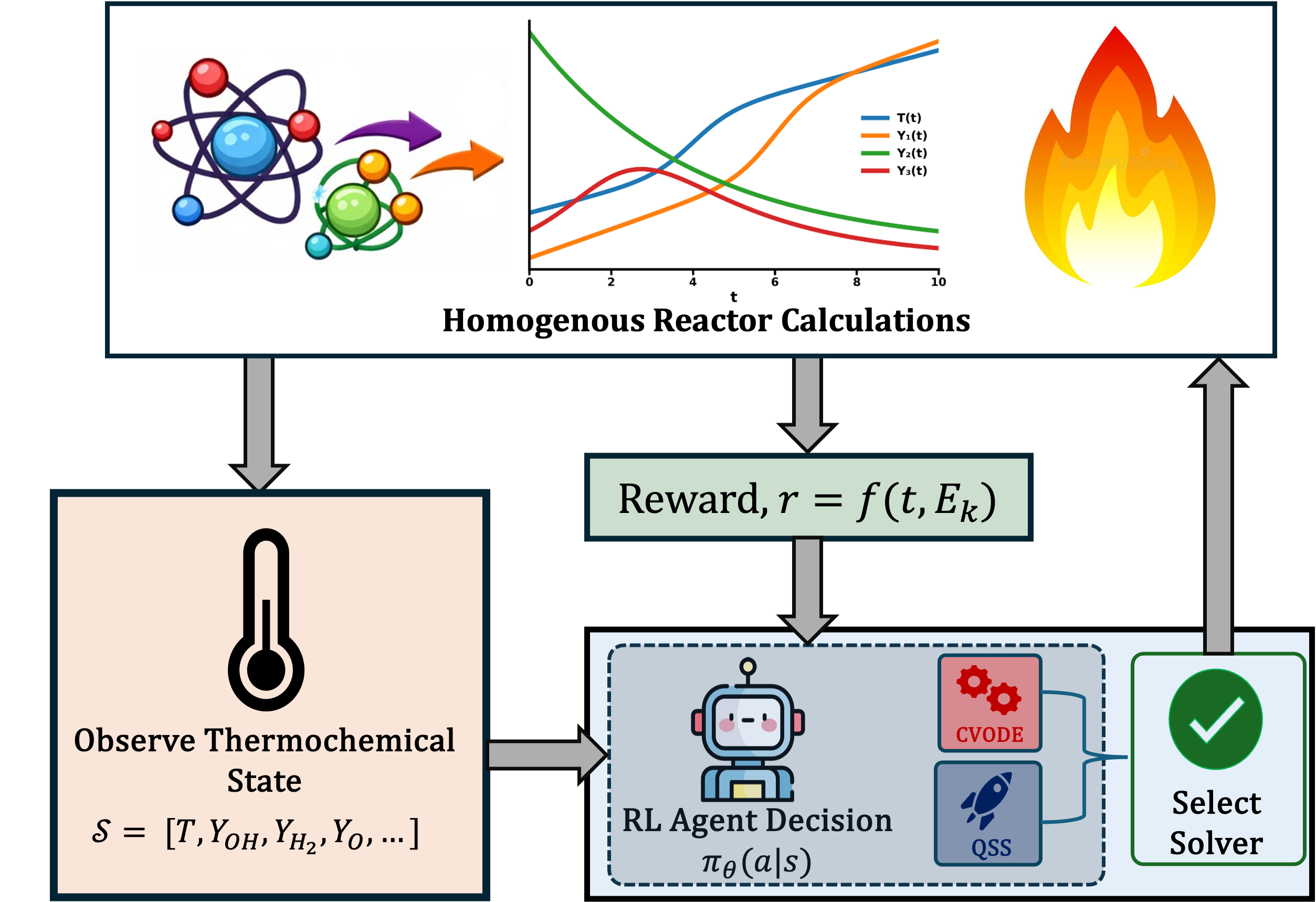}
    \caption{Reinforcement learning (RL) framework for adaptive solver selection. 
    The RL agent observes the evolving thermo-chemical state variables 
    (e.g., $T$, OH, H$_2$, O, etc.) from the combustion simulation and 
    selects the optimal numerical solver (e.g., CVODE or QSS) at each step.}
    \label{fig:adaptive_solver}
\end{figure*}

This section presents our reinforcement learning framework for adaptive solver selection in chemical kinetics integration illustrated in Fig. \ref{fig:adaptive_solver}). We begin by formulating the chemistry integration problem and describing the two solver classes, implicit backward differentiation formula (BDF) methods and quasi-steady-state (QSS) approximations. We then cast solver selection as a Markov Decision Process (MDP), detail our Lagrangian constraint-based reward formulation for balancing cost and accuracy, and describe the Proximal Policy Optimization (PPO) training procedure. Finally, we specify the training data generation and curriculum learning strategy.

\subsection{Chemistry Integration Problem}

Consider a reacting system at fixed or slowly varying pressure $p$. The evolution of $N_s$ chemical species and temperature is governed by a stiff system of ordinary differential equations (ODEs):
\begin{equation}
\frac{d\mathbf{y}}{dt} = \mathbf{f}(\mathbf{y}, T, p), 
\qquad
\frac{dT}{dt} = g(\mathbf{y}, T, p),
\label{eq:chem-ode}
\end{equation}
where $\mathbf{y}(t) \in \mathbb{R}^{N_s}$ denotes species mass fractions, $T(t)$ is temperature, $\mathbf{f}$ represents chemical production and destruction rates, and $g$, providing the temporal rate of change of system temperature, accounts for heat release from reactions. For typical hydrocarbon combustion, $N_s$ ranges from $\mathcal{O} (10)$ for reduced mechanisms of smaller hydrocarbon fuels to $\mathcal{O} (1,000)$ when dealing with detailed mechanisms for complex fuel molecules. Furthermore, the system exhibits extreme stiffness due to vastly different reaction timescales, identified mathematically by a Jacobian matrix $\mathbf{J} = \partial \mathbf{f}/\partial \mathbf{y}$ with eigenvalues that span many orders of magnitude, thereby creating numerical stability challenges for explicit integration methods.

\subsection{Solver Classes}

We consider two representative chemistry integrators that occupy opposite ends of the robustness-efficiency spectrum:

\subsubsection{CVODE: Implicit BDF Method}

CVODE from the SUNDIALS suite \cite{hindmarsh2005sundials} employs backward differentiation formulas (BDF), a family of implicit multistep methods. At each integration step from $t^n$ to $t^{n+1} = t^n + \Delta t$, BDF methods approximate the time derivative using a linear combination of solution values at previous time levels:
\begin{equation}
\frac{d\mathbf{y}}{dt}\bigg|_{t^{n+1}} \approx \frac{1}{\Delta t}\sum_{j=0}^{q} \alpha_j \mathbf{y}^{n+1-j},
\label{eq:bdf}
\end{equation}
where $q$ is the method order (typically 1--5, chosen adaptively) and $\{\alpha_j\}$ are BDF coefficients. Substituting \eqref{eq:bdf} into \eqref{eq:chem-ode} yields a nonlinear system for $\mathbf{y}^{n+1}$:
\begin{equation}
\mathbf{G}(\mathbf{y}^{n+1}) := \mathbf{y}^{n+1} - \sum_{j=1}^{q} \alpha_j \mathbf{y}^{n+1-j} - \Delta t \, \mathbf{f}(\mathbf{y}^{n+1}, T^{n+1}, p) = \mathbf{0}.
\label{eq:bdf-nonlinear}
\end{equation}
CVODE solves \eqref{eq:bdf-nonlinear} via Newton iteration:
\begin{equation}
\mathbf{y}^{n+1,(m+1)} = \mathbf{y}^{n+1,(m)} - \left[\mathbf{I} - \Delta t \frac{\partial \mathbf{f}}{\partial \mathbf{y}}\bigg|_{\mathbf{y}^{n+1,(m)}}\right]^{-1} \mathbf{G}(\mathbf{y}^{n+1,(m)}),
\label{eq:newton}
\end{equation}
where $m$ indexes the Newton iteration and the term in brackets is the iteration matrix. Each Newton step first requires a Jacobian evaluation that involves computing $\partial \mathbf{f}/\partial \mathbf{y}$ (either analytically or by finite differences), which amounts to an $\mathcal{O}(N_s^2)$ operation. Secondly, it involves the solution to a linear system solution by factorizing and solving the iteration matrix, typically $\mathcal{O}(N_s^3)$ for dense systems or $\mathcal{O}(N_s)$ for banded/sparse structures with iterative solvers. 

The implicit formulation allows CVODE to take relatively (compared to explicit methods) large stable time steps even when the system is stiff, since the method's stability region encompasses large portions of the negative real axis in the eigenvalue plane. However, the computational cost per step is substantial due to Jacobian evaluations and linear solves, particularly for detailed mechanisms where $N_s \gg 1$. 

\subsubsection{QSS: Quasi-Steady-State Approximation}

Quasi-steady-state (QSS) methods exploit the observation that in many reacting systems, a subset of species, typically short-lived radicals such as OH, O, H, HO$_2$ reach local chemical equilibrium much faster than the bulk species evolve \cite{mott2000quasi, Segel_Slemrod_1989, bond1998quasi, BOTHE2010120}. For these "fast" species indexed by $\mathcal{F} \subset \{1, \ldots, N_s\}$, we impose the algebraic constraint:
\begin{equation}
\mathbf{f}_{\mathcal{F}}(\mathbf{y}, T, p) = \mathbf{0},
\label{eq:qss-constraint}
\end{equation}
where $\mathbf{f}_{\mathcal{F}}$ denotes the production rate vector restricted to fast species. The remaining "slow" species $\mathcal{S} = \{1, \ldots, N_s\} - \mathcal{F}$ are integrated conventionally:
\begin{equation}
\frac{d\mathbf{y}_{\mathcal{S}}}{dt} = \mathbf{f}_{\mathcal{S}}(\mathbf{y}_{\mathcal{S}}, \mathbf{y}_{\mathcal{F}}, T, p),
\label{eq:qss-ode}
\end{equation}
with $\mathbf{y}_{\mathcal{F}}$ determined algebraically from \eqref{eq:qss-constraint} at each step. The coupled system \eqref{eq:qss-constraint}--\eqref{eq:qss-ode} can be advanced using explicit methods (e.g., Runge-Kutta) for the slow variables since the stiff fast modes have been eliminated algebraically.

In practice, the nonlinear system~\eqref{eq:qss-constraint} is solved iteratively (often via Newton's method or fixed-point iteration) within each time step. In this work, we employ the $\alpha$-QSS predictor-corrector scheme~\cite{mott2000quasi}, which splits each species equation into production ($q_i$) and destruction ($d_i$) terms:

\begin{equation}
\frac{dy_i}{dt} = q_i(\mathbf{y}, T) - d_i(\mathbf{y}, T),
\label{eq:qss-split}
\end{equation}
and employs a Padé approximation \cite{butcher2016rungekutta} to construct a second-order accurate update that transitions smoothly between explicit (non-stiff) and quasi-steady (stiff) regimes. The method uses an adaptive parameter $\alpha_i \in [0.5, 1]$ that interpolates between trapezoidal and backward Euler formulas based on the local stiffness ratio $\tau_i = d_i \Delta t / y_i$.

However, QSS can become inaccurate or unstable under certain conditions, such as when nearly all species evolve rapidly, or when the time step is too large relative to the fastest unconverged modes \cite{turanyi1993error}. 

\subsection{Adaptive Solver Selection as an MDP}

In this study, we present an approach to dynamically select CVODE or the $\alpha-$QSS solver at various regions of a thermochemical manifold, leveraging the strengths of both methods. We formulate solver selection as a sequential decision problem. At discrete decision points $t_k$ (typically aligned with the operator-splitting time step $\Delta t_{\text{global}}$ in CFD codes), an agent observes the local thermochemical state and chooses which solver to employ for advancing chemistry from $t_k$ to $t_{k+1} = t_k + \Delta t_k$ (see Figure \ref{fig:adaptive_solver}. The chosen solver integrates the chemistry equations, potentially taking multiple internal sub-steps, and returns updated states along with measured computational cost and error. This naturally defines a Markov Decision Process (MDP) consisting of a state space $\mathscr{S}$, an action space, $\mathcal{A}$, transition probability $P$, reward function $r$, and a discount factor $\gamma$: $(\mathscr{S}, \mathcal{A}, P, r, \gamma)$. These elements are discussed next.

\subsubsection{State Space $\mathscr{S}$}

The state $\mathbf{s}_k \in \mathscr{S}$ encodes local thermochemical information and lightweight contextual features sufficient for the agent to assess the local behavior of the system and select an appropriate time-integration scheme. Specifically, $\mathbf{s}_k$ is constructed as a feature vector comprising temperature, pressure, selected species mass fractions, and their local temporal rates of change. The temperature $T$ is normalized to have zero mean and unit variance, while pressure effects are incorporated through the logarithmic scaling $\log_{10}(p / p_{\text{atm}})$. To capture dominant reaction pathways and the evolution of the radical pool, we include mass fractions of key species,
\[
\mathbf{Y}_{\text{key}} =
[Y_{\text{OH}},\, Y_{\text{O}},\, Y_{\text{H}},\, Y_{\text{HO}_2},\,
 Y_{\text{H}_2\text{O}},\, Y_{\text{H}_2},\, Y_{\text{O}_2},\, Y_{\text{N}_2}],
\]
which involve both radical and stable species. Because species mass fractions vary over several orders of magnitude, each $Y_i$ is log-transformed as $\log_{10}(\max(Y_i, 10^{-20}))$ to accommodate a range of magnitudes between $10^{-20}$ and unity. In addition to instantaneous thermochemical quantities, the state includes temporal derivatives computed from the previous time step, namely the temperature rate of change $\Delta T / \Delta t$ and species gradient proxies $\Delta \mathbf{Y}_{\text{key}} / \Delta t$. These temporal gradients provide a direct and inexpensive proxy for stiffness: large magnitudes of $|\Delta T / \Delta t|$ or rapid species evolution signal fast transients associated with ignition, extinction, or flame-front dynamics. 

Overall, the state representation, as described, balances expressiveness with computational efficiency. Temperature and key species characterize the local thermochemical regime (e.g., lean versus rich, pre- versus post-ignition), while radicals such as OH and others are markers for rapid and high temperature heat release. By incorporating temporal gradients, the policy gains stiffness awareness without resorting to costly eigenvalue or Jacobian analyses. Importantly, we avoid including the full species vector ($N_s \gtrsim 100$) to keep inference overhead negligible. This enables solver selection decisions to be made cheaply and locally within large-scale reacting-flow simulations.

\subsubsection{Action Space $\mathcal{A}$}

The action set is discrete and binary:
\begin{equation}
\mathcal{A} = \{0: \texttt{CVODE}, \quad 1: \texttt{QSS}\}.
\label{eq:action-space}
\end{equation}
At decision point $t_k$, the agent selects $a_k \in \mathcal{A}$, and the environment invokes the corresponding solver to integrate chemistry from $t_k$ to $t_{k+1}$. We deliberately restrict to two actions to maximize the contrast between solver types and simplify policy interpretation. 

\subsubsection{Transition Dynamics $P$}

Given the current state $\mathbf{s}_k$ and selected action $a_k$, the environment advances the chemical system by executing the chosen chemistry integrator. This step produces the next state $\mathbf{s}_{k+1}$, which is constructed from the updated thermochemical fields $(\mathbf{y}^{n+1}, T^{n+1})$. In addition to the state transition, the environment returns two auxiliary quantities: the computational cost $C_k$, measured as the wall-clock time required for the chemistry kernel call (including Jacobian formation and linear solve overheads for implicit solvers such as CVODE), and an error estimate $E_k$, which quantifies the deviation from a high-accuracy reference solution as described later in Section~\ref{sec:error}. The resulting transition function $P(\mathbf{s}_{k+1} \mid \mathbf{s}_k, a_k)$ is deterministic in our setup, as chemical integration is fully determined by the current state and solver choice. Stochasticity arises only during training through randomized sampling of initial conditions.

\subsubsection{Reward Function $r$}
The reward function encodes the central trade-off in solver selection: reducing computational cost while enforcing a user-prescribed accuracy tolerance. We defer the full mathematical specification to Section~\ref{subsec:lagrangian-reward} after introducing the cost and error measures, but here provide a brief description of its role in the learning process.

In RL framework of this study, an agent is trained to maximize expected cumulative \textit{reward} through interaction with its environment (i.e., the chemical kinetics ODE solver). Consequently, the reward is the primary mechanism by which we shape the learned behavior: it incentivizes choosing the cheaper QSS update when it is accurate enough, and penalizes decisions that either incur unnecessary implicit-solver expense or allow errors to grow beyond acceptable levels through overuse of QSS. In this sense, the reward functions as an explicit control signal that shapes the learned switching policy toward efficient yet constraint-compliant time integration.



\subsection{Cost and Error Quantification}\label{sec:error}

\subsubsection{Computational Cost $C_k$}

We measure the wall-clock time required by the chemistry integrator to advance from $t_k$ to $t_{k+1}$:
\begin{equation}
C_k = t_{\text{wall}}^{\text{end}} - t_{\text{wall}}^{\text{start}},
\label{eq:cost-raw}
\end{equation}
This includes all overhead: function evaluations, Jacobian assembly, linear solves, convergence checks, and error control. To stabilize training, we normalize costs relative to a baseline:
\begin{equation}
\widetilde{C}_k = \frac{C_k}{\bar{C}_{\text{CVODE}}},
\label{eq:cost-normalized}
\end{equation}
where $\bar{C}_{\text{CVODE}}$ is the mean CVODE cost per decision window for the same thermochemical trajectory (precomputed). This normalization ensures $\widetilde{C}_k \approx 1$ for CVODE and $\widetilde{C}_k \ll 1$ for QSS in favorable regimes, making the reward magnitudes comparable across different test cases.
 
\subsubsection{Solution Error $E_k$}

We quantify the accuracy of the selected solver by comparing its output to a high-fidelity reference solution. Let $(\mathbf{y}_k^{\text{sel}}, T_k^{\text{sel}})$ denote the state produced by the selected solver at $t_k$, and $(\mathbf{y}_k^{\text{ref}}, T_k^{\text{ref}})$ denote the reference solution. We define a composite error metric:
\begin{equation}
E_k = \alpha_T \frac{\|T_k^{\text{sel}} - T_k^{\text{ref}}\|}{\|T_k^{\text{ref}}\| + \varepsilon_T}
+ \alpha_Y \frac{\|\mathbf{Y}_k^{\text{sel}} - \mathbf{Y}_k^{\text{ref}}\|}{\|\mathbf{Y}_k^{\text{ref}}\| + \varepsilon_Y},
\label{eq:error}
\end{equation}

where $\|\cdot\|_2$ denotes the Euclidean norm, $\alpha_T + \alpha_Y = 1$ are user-defined weights ($\alpha_T = 0.7$, $\alpha_Y = 0.3$), and $\varepsilon_T= \varepsilon_Y = 10^{-12}$ prevent division by zero.
We precompute CVODE trajectories with tight tolerances (rtol = $10^{-10}$, atol = $10^{-20}$) for all training cases. These serve as ground truth throughout training.

\subsection{Lagrangian Constraint Reward Formulation}
\label{subsec:lagrangian-reward}

We seek solver-selection policies that reduce computational cost while satisfying a user-prescribed accuracy tolerance $\epsilon$. To encode this requirement in a form amenable to constrained reinforcement learning, we define a per-step constraint violation signal
\begin{equation}
\nu_k \;\equiv\; \max(0,\,E_k-\epsilon),
\label{eq:hinge_violation}
\end{equation}
which measures the amount by which the local integration error $E_k$ exceeds the tolerance. Our objective is then to minimize the expected computational cost while constraining the expected average violation:
\begin{equation}
\min_{\pi} \; \mathbb{E}_{\pi}\!\left[\sum_{k=0}^{K-1} \widetilde{C}_k \right]
\quad \text{subject to} \quad
\mathbb{E}_{\pi}\!\left[\frac{1}{K} \sum_{k=0}^{K-1} \nu_k \right] \le \delta,
\label{eq:constrained-objective}
\end{equation}
where $\pi$ denotes the policy, $K$ is the episode length, and the expectation is taken over trajectories induced by $\pi$ from the initial condition distribution. Here, $\delta \ge 0$ specifies the allowable average constraint violation (with $\delta=0$ corresponding to a strict no-violation target). This defines a constrained Markov decision process (CMDP) \cite{Altman2021-ls}, which we handle using a primal-dual Lagrangian formulation.

\subsubsection{Primal-Dual Lagrangian Approach}

We introduce a Lagrange multiplier $\lambda \ge 0$ associated with the expected violation constraint in eqn \eqref{eq:constrained-objective}. Using the violation signal $\nu_k$ defined in eqn \eqref{eq:hinge_violation}, the Lagrangian is,
\begin{equation}
\mathcal{L}(\pi,\lambda)
=
\mathbb{E}_{\pi}\!\left[\sum_{k=0}^{K-1}\widetilde{C}_k\right]
+
\lambda\left(
\mathbb{E}_{\pi}\!\left[\frac{1}{K}\sum_{k=0}^{K-1}\nu_k\right]-\delta
\right),
\qquad \lambda \ge 0,
\label{eq:lagrangian}
\end{equation}
which yields the dual problem $\max_{\lambda \ge 0}\min_{\pi}\mathcal{L}(\pi,\lambda)$. We solve this using alternating (primal-dual) updates:
\begin{enumerate}
    \item \textbf{Policy update (primal):} for a fixed $\lambda$, optimize $\pi$ (via PPO) to minimize the penalized objective implied by \eqref{eq:lagrangian}.
    \item \textbf{Multiplier update (dual):} update $\lambda$ based on the observed average violation $\frac{1}{K}\sum_{k=0}^{K-1}\nu_k$ to drive the constraint toward satisfaction (see Section~\ref{dual-variable-update}).
\end{enumerate}

\subsubsection{Step-wise Reward}\label{sec:reward}

From eqn \eqref{eq:lagrangian}, we obtain a step-wise reward by combining the instantaneous computational cost with a penalty on accuracy violations. In practice, we implement solver selection via a step-wise reward that penalizes accuracy violations in proportion to the multiplier:
\begin{equation}
r_k = -\widetilde{C}_k - \lambda\,\nu_k .
\label{eq:reward-main}
\end{equation}
Here, $\nu_k$ penalizes exceedance of the prescribed tolerance $\epsilon$, while $\lambda \ge 0$ controls the strength of constraint enforcement. The reward \eqref{eq:reward-main} implements a soft constraint on accuracy. When $E_k \le \epsilon$, the violation is zero ($\nu_k = 0$), so $r_k = -\widetilde{C}_k$ and the agent is incentivized to minimize computational cost. When $E_k > \epsilon$, the violation becomes $\nu_k = E_k - \epsilon > 0$, leading to an additional penalty $-\lambda(E_k-\epsilon)$ that discourages solver choices producing excessive error. Note that eqn \eqref{eq:reward-main} corresponds to the per-step penalized objective induced by eqn \eqref{eq:lagrangian} up to a constant rescaling of the multiplier (i.e., absorbing the factor $1/K$ into $\lambda$), which does not change the form of the learned policy update but simplifies implementation.

As $\lambda$ adapts (see Section~\ref{dual-variable-update}), the policy is driven toward solutions that keep the average violation near the allowable level $\delta$ while minimizing cost, thereby learning an accuracy-efficiency trade-off consistent with the CMDP formulation. In this work, we set $\delta=0$.

\subsubsection{Dual Variable Update}
\label{dual-variable-update}

At the end of each training episode (or every $M$ episodes), we update $\lambda$ via projected gradient ascent on the dual objective:
\begin{equation}
\lambda \leftarrow \text{clip}\left(\lambda + \eta_\lambda \left(\overline{\nu} - \delta\right), 0, \lambda_{\max}\right),
\label{eq:dual-update}
\end{equation}
where $\overline{\nu} = \frac{1}{K} \sum_{k=0}^{K-1} \nu_k$ is the mean constraint violation over the episode, $\eta_\lambda$ is the dual learning rate ($10^{-3}$), and $\delta \ge 0$ is an allowance ($10^{-4}$) that permits the policy to slightly exceed $\epsilon$ without penalty. On the other hand, $\lambda_{\max}$ prevents unbounded growth ($10^2$).
Therefore, if $\overline{\nu} > \delta$ (i.e., policy violates the constraint), $\lambda$ increases, tightening the penalty and encouraging the policy to prioritize accuracy. Conversely, if $\overline{\nu} < \delta$ (policy satisfies the constraint), $\lambda$ decreases, relaxing the penalty and allowing more aggressive computational cost reduction.

\subsection{Policy Architecture and Training}

\subsubsection{Actor-Critic Neural Network}

We employ a standard actor-critic architecture with a shared feature set. Let $\phi_\theta: \mathcal{S} \to \mathbb{R}^{d_h}$ denote a multilayer perceptron (MLP) encoder with parameters $\theta$, mapping the state $s$ to a hidden representation of dimension $d_h$. The encoder consists of two fully connected layers with hyperbolic tangent activations, each with 128 neurons.
Orthogonal weight initialization \cite{saxe2013exact} is used to mitigate gradient vanishing.

The \textbf{actor} (policy) head outputs a categorical distribution over actions:
\begin{equation}
\pi_\theta(a | s) = \text{Softmax}(W_\pi \phi_\theta(s)),
\label{eq:actor}
\end{equation}
where $W_\pi \in \mathbb{R}^{|\mathcal{A}| \times d_h}$ is the actor weight matrix. For our binary action space, $\pi_\theta$ produces two logits that are normalized via softmax. The \textbf{critic} (value function) head predicts the expected cumulative reward:
\begin{equation}
V_\theta(s) = w_v^\top \phi_\theta(s),
\label{eq:critic}
\end{equation}
where $w_v \in \mathbb{R}^{d_h}$ is the critic weight vector.

\subsubsection{Proximal Policy Optimization (PPO)}

We train the policy using Proximal Policy Optimization \cite{schulman2017proximal}, a policy gradient method that balances sample efficiency and stability via clipped surrogate objectives. PPO collects trajectories using the current policy $\pi_{\theta_{\text{old}}}$, computes advantages via Generalized Advantage Estimation (GAE) \cite{schulman2015high}, and updates $\theta$ by maximizing a clipped objective that prevents excessively large policy updates.

\paragraph{Advantage estimation.} For each timestep $k$ in a trajectory, we compute the GAE advantage:
\begin{equation}
\begin{split}
\hat{A}_k &= \sum_{l=0}^{K-k-1} (\gamma \lambda_{\text{GAE}})^l \delta_{k+l}, \\
\text{where} \quad
\delta_k &= r_k + \gamma V_{\theta_{\text{old}}}(s_{k+1}) - V_{\theta_{\text{old}}}(s_k),
\end{split}
\label{eq:gae}
\end{equation}

with $\gamma = 0.99$ (discount factor) and $\lambda_{\text{GAE}} = 0.95$ (GAE parameter). The advantages are normalized to zero mean and unit variance across the minibatch to stabilize training.

\paragraph{Loss function.} The PPO loss combines three terms:
\begin{align}
\mathcal{L}_{\text{PPO}}(\theta) &= \mathcal{L}_{\text{policy}}(\theta) + \beta_V \mathcal{L}_{\text{value}}(\theta) - \beta_H \mathcal{L}_{\text{entropy}}(\theta),
\label{eq:ppo-loss}
\end{align}
where $\mathcal{L}_{\text{policy}}(\theta)$ is the policy loss involving a clipped surrogate, $\mathcal{L}_{\text{value}}(\theta)$ is the value loss, and $\mathcal{L}_{\text{entropy}}(\theta)$ is the entropy loss.

Coefficients are set to $\beta_V = 0.5$ (value loss weight) and $\beta_H = 0.01$ (entropy bonus) as used in \cite{schulman2017proximal}. The entropy term encourages exploration early in training; as the policy converges, $\mathcal{H}(\pi_\theta)$ naturally decreases.

\paragraph{Training procedure.} For each policy update, the following steps are performed:
\begin{enumerate}
    \item Collect $N_{\text{rollout}} = 2048$ timesteps of experience using $\pi_{\theta_{\text{old}}}$.
    \item Compute advantages $\{\hat{A}_k\}$ and returns $\{\hat{V}_k\}$ via GAE.
    \item Perform $N_{\text{epoch}} = 4$ epochs of minibatch gradient descent on $\mathcal{L}_{\text{PPO}}$ with batch size 64.
    \item Update $\theta$ via Adam optimizer \cite{kingma2014adam} with learning rate $3 \times 10^{-4}$.
\end{enumerate}

\subsection{Training Data}

\subsubsection{0D Homogeneous Reactors}

Training episodes consist of constant-pressure homogeneous ignition simulations with selected values of initial temperatures, pressure, and mixture fractions chosen as $T_0 \in [300 \text{ K}, 1200 \text{ K}]$ (uniform sampling), $p \in [1 \text{ atm}, 60 \text{ atm}]$ (log-uniform sampling) and $Z \in [10^{-6}, 10^{-1}]$ (log-uniform), respectively. For training, we use a detailed n-dodecane mechanism with 106 species and 678 reactions \cite{luo2014development}. 

At each sampled condition, we precompute a CVODE reference trajectory with tolerances $\mathrm{rtol} = 10^{-10}$, $\mathrm{atol} = 10^{-20}$, and integrate to $t_{\text{final}} = 10\,\mathrm{ms}$ (or until temperature exceeds $2500\,\mathrm{K}$). Decision points $\{t_k\}$ are spaced at $\Delta t = 1\,\mu\mathrm{s}$ (for 0D) or aligned with the CFD global timestep (in 1D). This yields $K \sim 100$--$10{,}000$ decisions per episode depending on ignition delay.

All training experiments were executed on a laptop-class workstation (Apple MacBook Pro with an M2 processor and 16~GB unified memory). Under this setup, training for $10^5$ environment steps completes in under three hours, including on-the-fly evaluation of per-step cost and error metrics.
Key hyperparameters used for training are summarized in Table~\ref{tab:hyperparameters}.

\begin{table}[h]
\centering
\caption{Summary of key hyperparameters}
\label{tab:hyperparameters}
\begin{tabular}{ll}
\hline
\textbf{Parameter} & \textbf{Value} \\
\hline
Rollout length $N_{\text{rollout}}$ & 2048 \\
Minibatch size & 64 \\
PPO epochs $N_{\text{epoch}}$ & 4 \\
Learning rate & $3 \times 10^{-4}$ \\
Discount $\gamma$ & 0.99 \\
GAE $\lambda_{\text{GAE}}$ & 0.95 \\
Clip threshold $\epsilon_{\text{clip}}$ & 0.2 \\
Entropy coefficient $\beta_H$ & 0.01 \\
Value coefficient $\beta_V$ & 0.5 \\
Dual learning rate $\eta_\lambda$ & $10^{-3}$ \\
Constraint tolerance $\epsilon$ & $10^{-3}$ \\
Initial $\lambda$ & 1.0 \\
$\lambda_{\max}$ & 100 \\
\hline
\end{tabular}
\end{table}

\section{Results and Discussion}
\label{sec:results}

While the policy was trained exclusively on 0D trajectories based on initial conditions described in the preceding section, to demonstrate some degree of generalizability, we evaluate the learned RL-adaptive solver selection policy on both zero-dimensional (0D) homogeneous reactors and one-dimensional (1D) counterflow diffusion flames. We assess performance through three key metrics: (i) computational speedup relative to CVODE-only baseline, (ii) solution accuracy measured by temperature and ignition delay errors, and (iii) solver utilization patterns that reveal the policy's learned strategy. 

\subsection{Zero-Dimensional Homogeneous Reactors}
\label{subsec:0d_results}

\subsubsection{Computational Performance and Speedup}

\begin{figure*}[htbp]
    \centering
    \begin{subfigure}[b]{0.48\textwidth}
        \includegraphics[width=\textwidth]{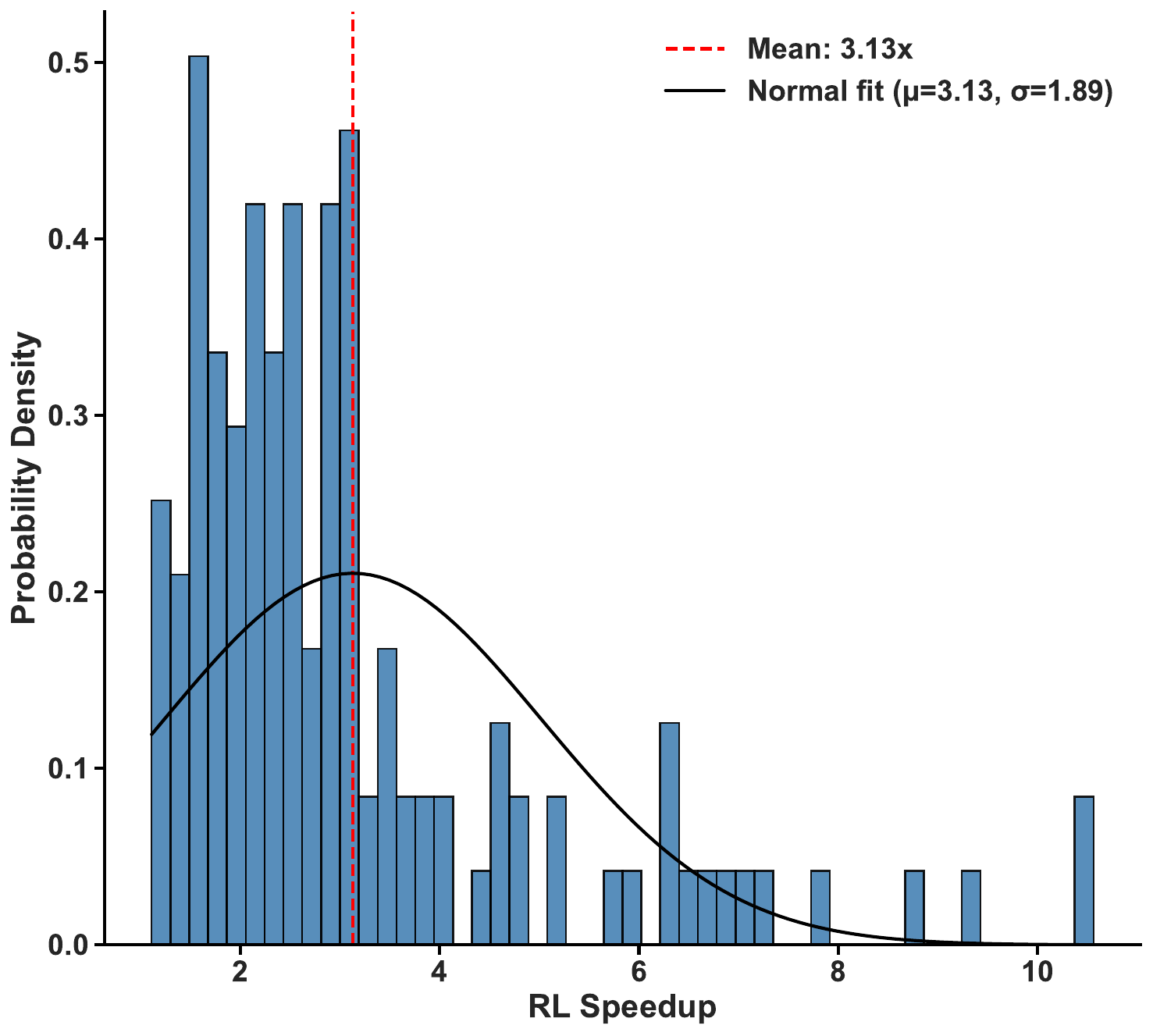}
        \caption{Speedup distribution over 0D test cases. The RL-adaptive policy achieves $3.13\times$ mean speedup over CVODE ($\sigma=1.89$). Black: normal fit; red dashed: mean.}
        \label{fig:speedup_dist}
    \end{subfigure}
    \hfill
    \begin{subfigure}[b]{0.48\textwidth}
        \includegraphics[width=\textwidth]{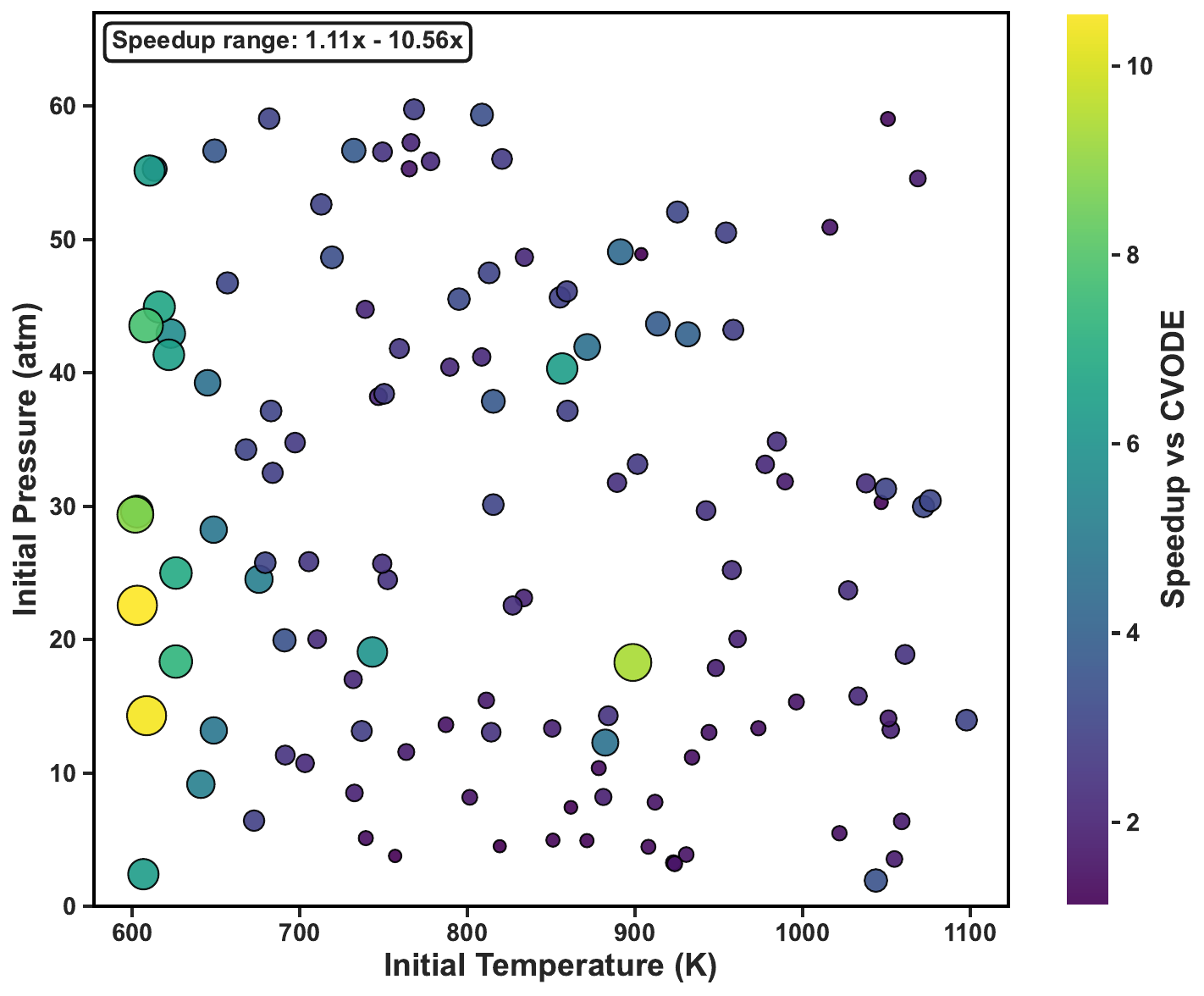}
        \caption{Speedup versus initial conditions. Markers denote cases (color: speedup; size: \% CVODE usage). Speedup ranges from $1.1\times$ to $10.5\times$ over $T_0\in[600,1100]~\mathrm{K}$ and $p\in[1,60]~\mathrm{atm}$.}
        \label{fig:speedup_vs_conditions}
    \end{subfigure}
    
    \caption{RL-adaptive performance across initial conditions: (a) speedup distribution and (b) dependence on $(T_0,p)$. Accuracy is maintained within $2.6\%$ ignition-delay error and <$110~\mathrm{K}$ temperature RMSE.}
    \label{fig:rl_speedup_results}
\end{figure*}

Figure~\ref{fig:rl_speedup_results} summarizes the speedups achieved by the RL-adaptive policy across the sampled 0D conditions. Figure~\ref{fig:rl_speedup_results}a shows a histogram of the resulting speedup distribution, with a mean speedup of approximately $3\times$, a minimum of $1.11\times$, and a maximum exceeding $10\times$. Figure~\ref{fig:rl_speedup_results}b maps these speedups in the $(T_0,p)$ space, where marker size scales with speedup. Several trends are apparent. The largest efficiency gains are generally realized at lower initial temperatures across the full pressure range, with a modest reduction in speedup as pressure increases. As $T_0$ increases, the gains diminish, particularly at lower pressures, consistent with faster ignition kinetics and more sustained stiffness which forces the RL agent to employ the more expensive CVODE solver more frequently. While a few isolated high-speedup points appear at higher $T_0$, these are likely attributable to stochastic variability rather than a robust trend.

Table~\ref{tab:0d_results} reports representative accuracy and efficiency metrics at selected thermodynamic conditions. The maximum speedup among these representative cases occurs at the lowest-temperature, lowest-pressure case ($T_0=650~\mathrm{K}$, $p=1~\mathrm{atm}$), where the long induction period ($97.2~\mathrm{ms}$ ignition delay) creates an extended window of slow chemical evolution during which QSS integration is sufficient. In this regime, the policy selects CVODE for only $5.4\%$ of time steps, concentrating expensive implicit integration near the ignition event (Figure~\ref{fig:0d_cond1}, left panel; red markers at $t \approx 15$--$22~\mathrm{ms}$). In contrast, the $(T_0,p)=(1000~\mathrm{K},30~\mathrm{atm})$ case exhibits the smallest speedup ($1.42\times$) and the highest CVODE usage ($68.0\%$). Here, rapid ignition kinetics ($1.40~\mathrm{ms}$) and steep transients over a larger fraction of the trajectory compel frequent implicit integration to maintain accuracy. Even in this more stringent regime, the policy reduces total computational cost by $29.6\%$ relative to pure implicit integration, demonstrating adaptive allocation of computational effort under strongly stiff conditions.

Finally, policy inference overhead remains negligible across all conditions, averaging $67.9~\mathrm{ms}$ (approximately $1\%$ of total CPU time). This confirms that neural network evaluation (implemented via batched forward passes) introduces minimal additional cost, supporting the feasibility of the approach in production CFD workflows where chemistry integration dominates runtime.

\begin{table*}[htbp]
\centering
\caption{Performance metrics for RL-adaptive solver selection in 0D homogeneous reactors compared to CVODE and QSS baselines. All cases use n-dodecane mechanism with 106 species.}
\label{tab:0d_results}
\begin{adjustbox}{width=\textwidth}
\begin{tabular}{l|ccc|c|ccc|cc}
\hline
& \multicolumn{4}{c|}{\textbf{Ignition Delay}} 
& \multicolumn{3}{c|}{\textbf{CPU Time}} 
& \multicolumn{2}{c}{\textbf{Performance}} \\
\cline{2-10}
\textbf{Condition} 
& \textbf{CVODE} & \textbf{QSS} & \textbf{RL} & \textbf{Error} 
& \textbf{CVODE} & \textbf{QSS} & \textbf{RL} 
& \textbf{Speedup} & \textbf{Inference} \\
\textbf{$(T_0, P)$}
& \textbf{(ms)} & \textbf{(ms)} & \textbf{(ms)} & \textbf{(\%)} 
& \textbf{(s)} & \textbf{(s)} & \textbf{(s)} 
& \textbf{vs CVODE} & \textbf{(s)} \\
\hline
650K, 1.0atm & 97.21 & 109.94 & 96.81 & 0.41 & 39.25 & 10.12 & 12.06 & 3.25$\times$ & 0.11 \\
800K, 10.0atm & 2.30 & 2.42 & 2.27 & 1.13 & 12.92 & 0.74 & 6.43 & 2.01$\times$ & 0.07 \\
1000K, 30.0atm & 1.40 & 1.64 & 1.41 & 0.64 & 4.92 & 1.58 & 3.95 & 1.24$\times$ & 0.02 \\
750K, 60.0atm & 1.45 & 2.10 & 1.49 & 2.62 & 6.92 & 0.48 & 4.30 & 1.61$\times$ & 0.07 \\
\hline
\end{tabular}
\end{adjustbox}
\end{table*}

\subsubsection{Solution Accuracy and Learned Solver-Selection Strategy}

\begin{figure*}[t]
    \centering
    
    \begin{subfigure}[b]{0.48\textwidth}
        \includegraphics[width=\textwidth]{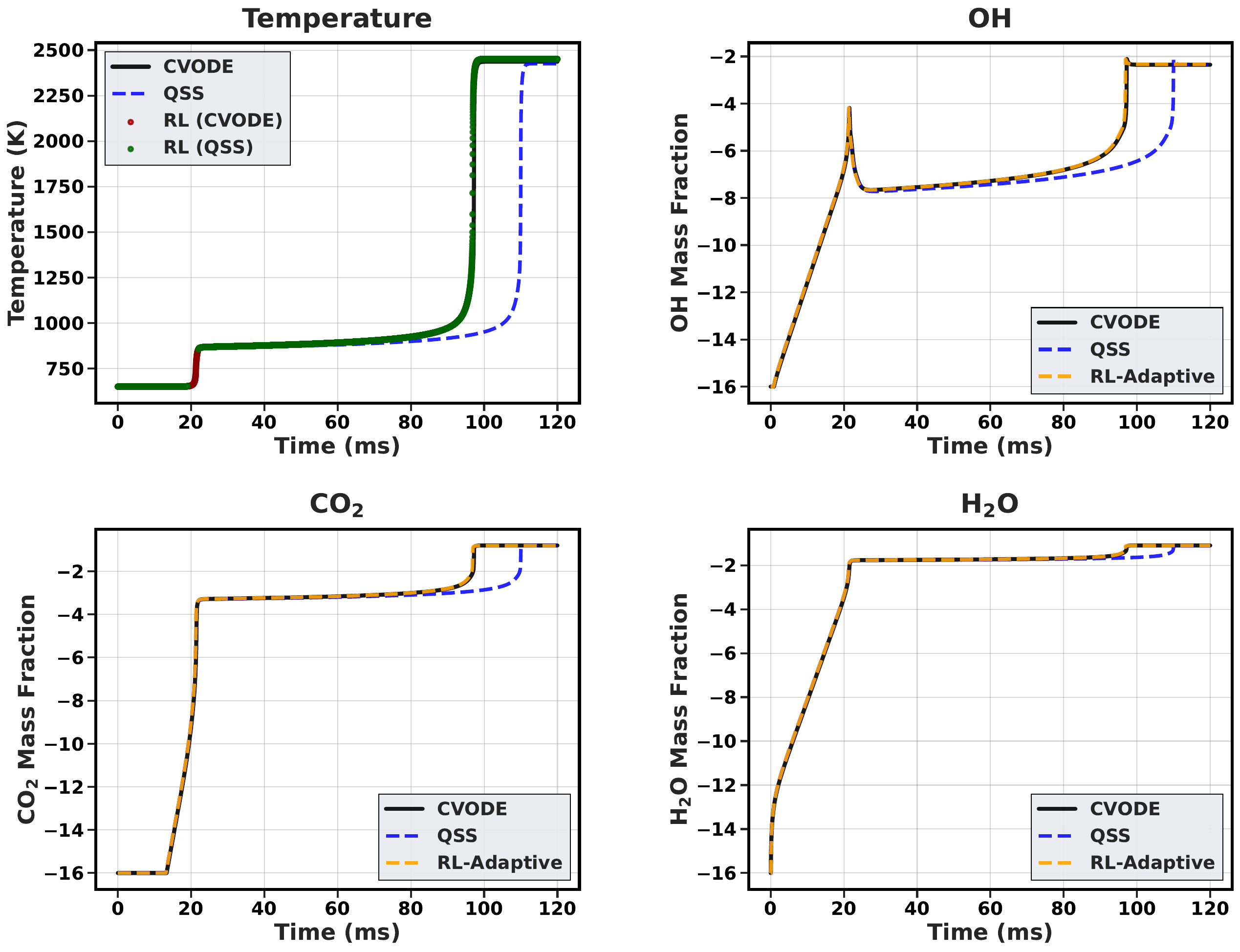}
        \caption{Condition 1: 650K, 1.0atm}
        \label{fig:0d_cond1}
    \end{subfigure}
    \hfill
    \begin{subfigure}[b]{0.48\textwidth}
        \includegraphics[width=\textwidth]{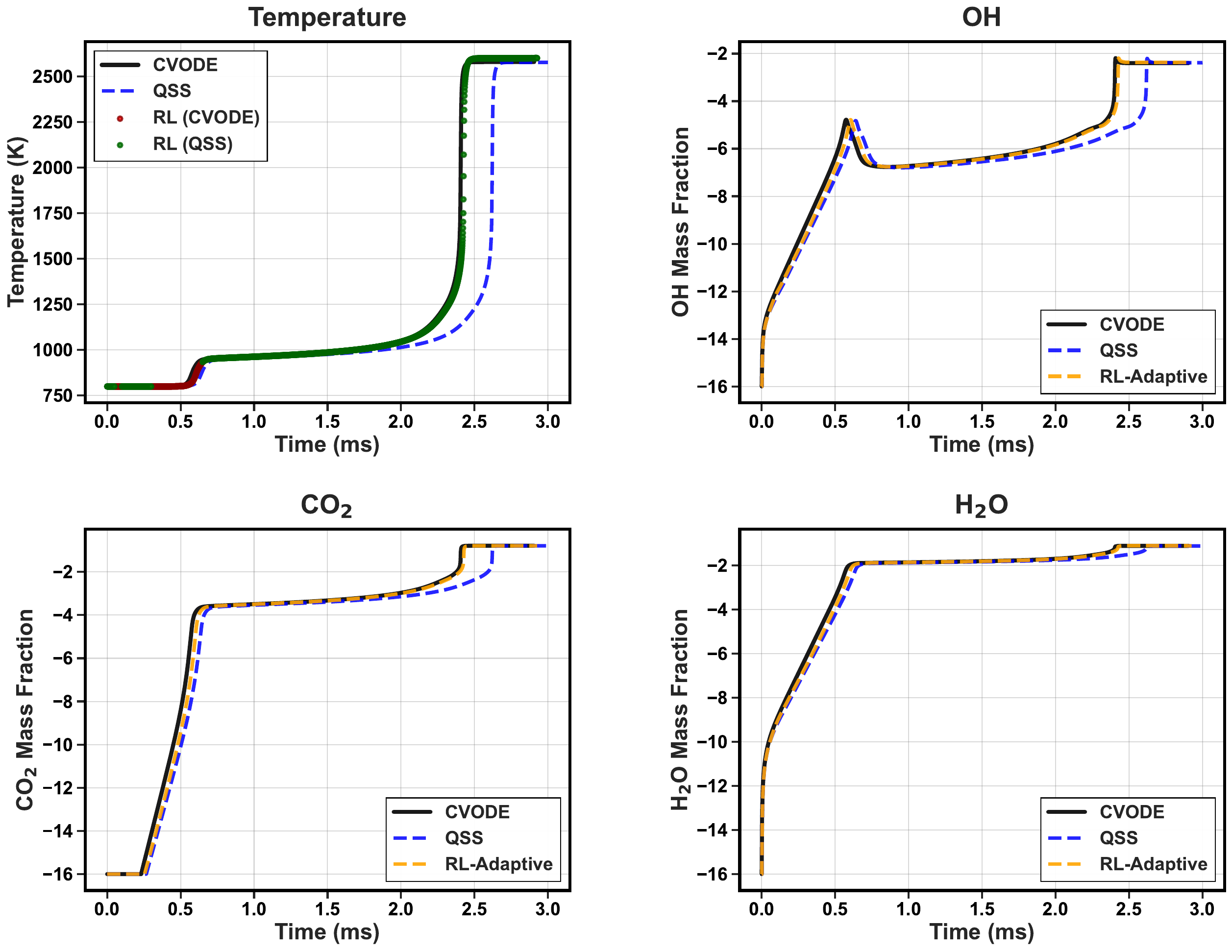}
        \caption{Condition 2: 800K, 10.0atm}
        \label{fig:0d_cond2}
    \end{subfigure}
    
    \vspace{0.5cm}
    
    \begin{subfigure}[b]{0.48\textwidth}
        \includegraphics[width=\textwidth]{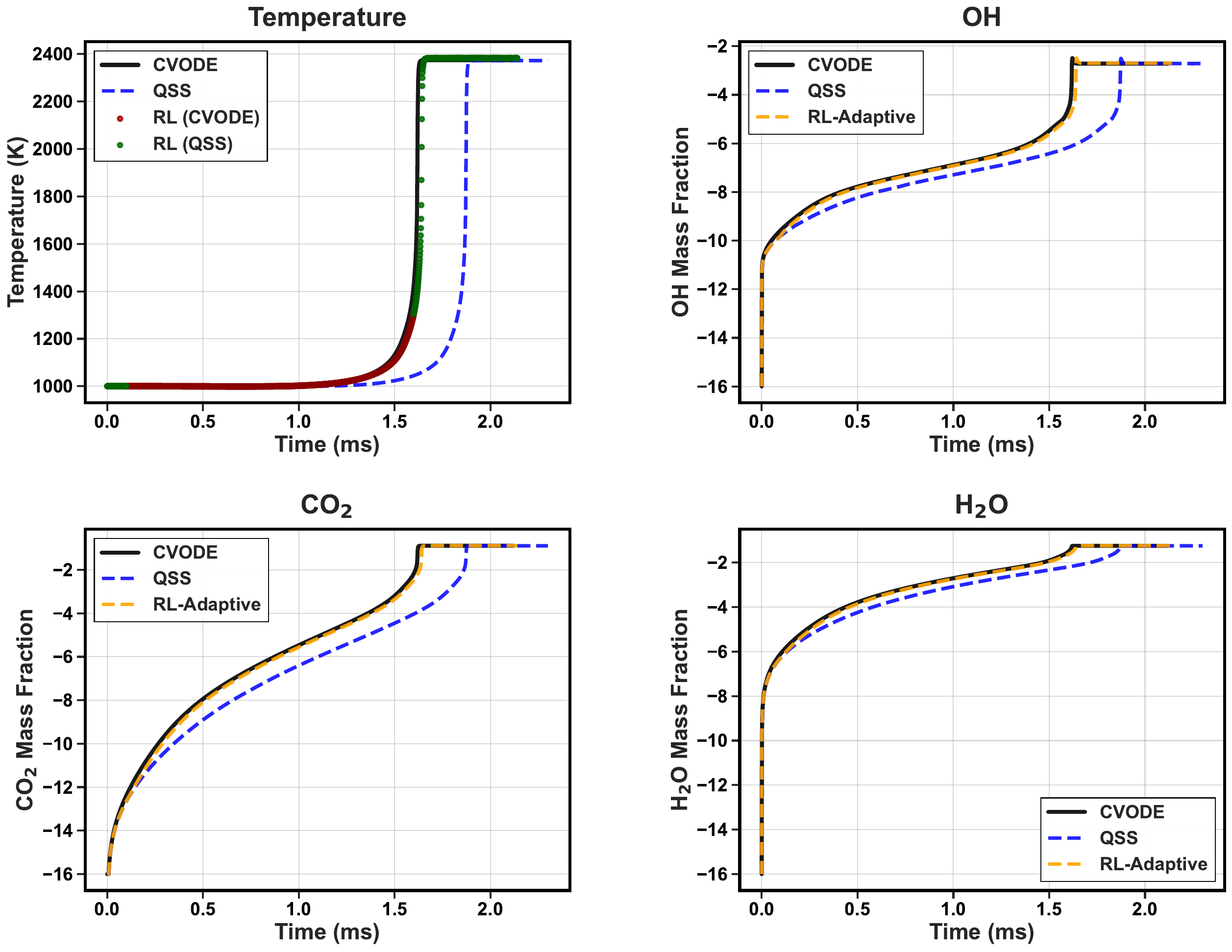}
        \caption{Condition 3: 1000K, 30.0atm}
        \label{fig:0d_cond3}
    \end{subfigure}
    \hfill
    \begin{subfigure}[b]{0.48\textwidth}
        \includegraphics[width=\textwidth]{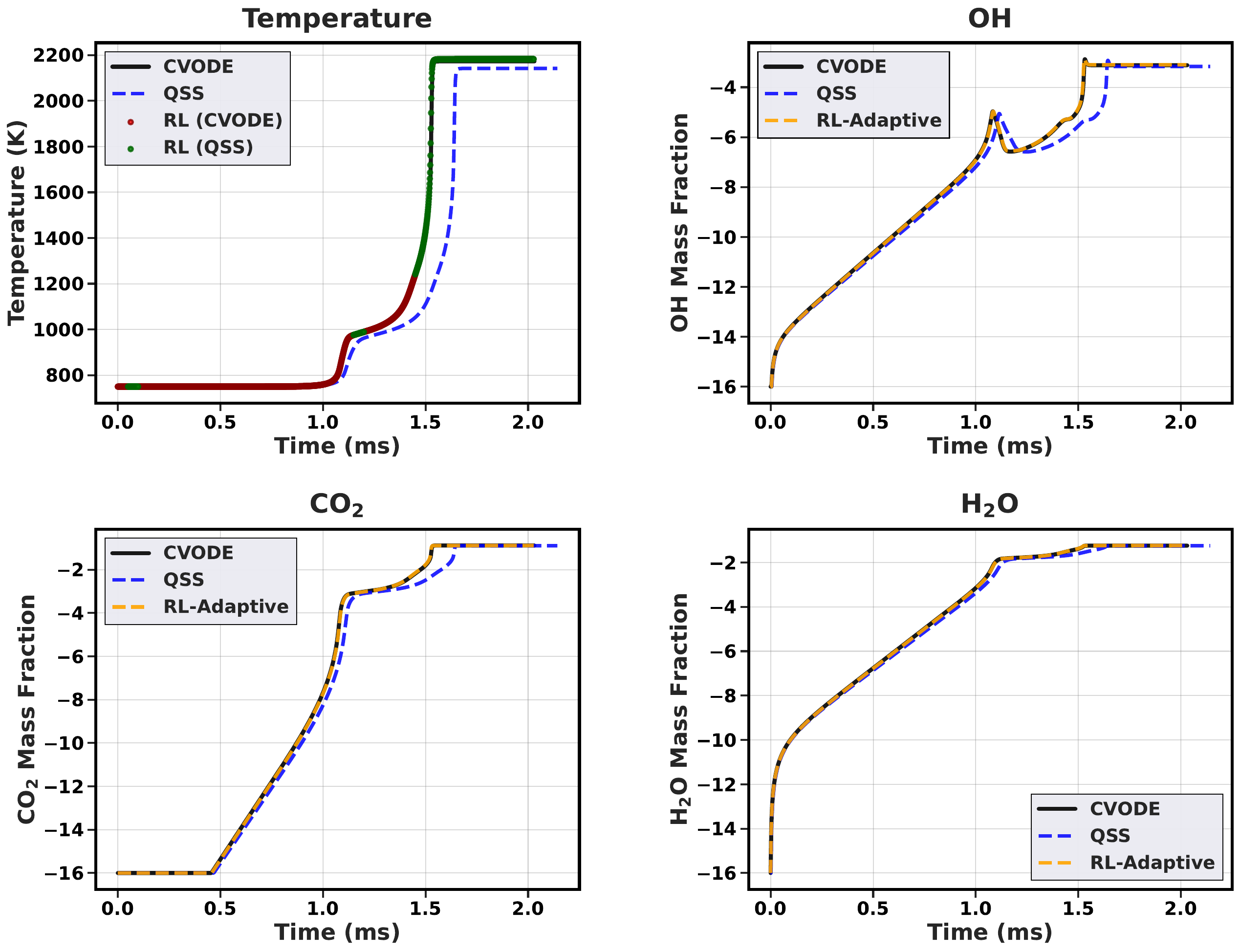}
        \caption{Condition 4: 750K, 60.0atm}
        \label{fig:0d_cond4}
    \end{subfigure}
    
    \caption{0D trajectory comparisons across thermochemical conditions using the RL-adaptive solver policy. 
    \textbf{(a)} Condition 1 (650K, 1.0atm) achieves a 3.25$\times$ speedup with 0.41\% ignition delay error and 5.4\% CVODE usage. The temperature subplot shows solver selection via colored scatter (red=CVODE, green=QSS), with CVODE concentrated in a narrow ignition window ($t \approx 15$--22~ms). 
    \textbf{(b--d)} As temperature and pressure increase (hence stiffness), CVODE usage rises from 14\% (b) to 68.0\% (c,d), demonstrating adaptive resource allocation. In all cases, the RL-adaptive approach maintains ignition delay errors below 2.62\% and temperature RMSE below 108K.}
    
    \label{fig:0d_trajectories}
\end{figure*}

Across all test cases, the learned policy satisfies the prescribed accuracy requirement ($\epsilon = 10^{-3}$), maintaining ignition-delay errors below $2.62\%$ (Table~\ref{tab:0d_results}). Post-ignition equilibrium temperatures agree with the reference solution to within $0.1\%$, and key species profiles (OH, CO, and H$_2$O) exhibit root-mean-square (RMS) errors below $10^{-2}$ in mass fraction.

Figure~\ref{fig:0d_cond1} illustrates the trajectory for the low-temperature, low-pressure case (Condition~1: $T_0=650~\mathrm{K}$, $p=1~\mathrm{atm}$). The RL-adaptive solution closely follows the CVODE reference throughout both the low- and high-temperature stages of ignition, whereas QSS overpredicts the second-stage ignition delay by $13.1\%$. While QSS adequately captures the initial low-temperature chemistry, errors introduced during this stage accumulate and degrade accuracy during the high-temperature ignition phase. In contrast, the RL policy selectively invokes CVODE at critical portions of the trajectory where QSS accuracy degrades, preserving accuracy across both stages while retaining some of the improved computational savings of QSS.

At intermediate initial temperature and pressure (Fig.~\ref{fig:0d_cond2}, $T_0 = 800~\mathrm{K}$, $p = 10~\mathrm{atm}$), a similar two-stage ignition behavior is observed, with QSS again overpredicting the second-stage ignition delay. In this case, however, error accumulation is already evident during the low-temperature first-stage ignition, as reflected in deviations in the temporal profiles of intermediate species such as \ce{CO2}, OH, and \ce{H2O}. Interestingly, the RL-adaptive solver switches to CVODE precisely in this region, where QSS-induced errors begin to accumulate and the solution starts to diverge from the reference. After stabilizing the solution through implicit integration, the policy reverts to QSS for the remainder of the trajectory, while still maintaining close agreement with the CVODE solution.

This behavior suggests that for QSS, the dominant errors observed during the second-stage ignition originate from inaccuracies introduced earlier in the first-stage chemistry. It also demonstrates a key advantage of the RL-based approach over static solver-selection strategies based on neural classifiers trained in supervised mode using labels: by leveraging experience acquired during training, the RL agent is able to effectively account for potential error accumulation in future time steps. To mitigate these future unseen errors, it adapts solver choices to prevent error accumulation before it degrades solution fidelity.

Condition~3 (Fig.~\ref{fig:0d_cond3}, $T_0 = 1000~\mathrm{K}$, $p = 30~\mathrm{atm}$) involves single-stage ignition and exhibits more pervasive CVODE usage. In this regime, highly reactive conditions that are challenging to the QSS solver emerge early in the trajectory, causing QSS accuracy to degrade shortly after the induction period begins. The RL agent correctly identifies this behavior (as well as the downstream effects) and correspondingly invokes CVODE over a larger fraction of the simulation to preserve solution fidelity. At a selected high-pressure condition (Fig.~\ref{fig:0d_cond4}, $T_0 = 750~\mathrm{K}$, $p = 60~\mathrm{atm}$), the agent uses CVODE for most of the trajectory. In this case, a cursory evaluation of the profiles suggest that QSS may be sufficient for much of the early induction period, and it is unclear whether the conservative behavior of the RL agent reflects strong sensitivity of later ignition dynamics to early-stage accuracy, or residual suboptimality arising from training stochasticity or data balance issues. Nevertheless, as a general observation, the RL agent learns to select CVODE when required to maintain accuracy, while switching to QSS for improved computational efficiency when feasible.

Overall, the learned solver-selection strategy generalizes beyond fixed heuristic rules that either overuse implicit integration (sacrificing efficiency) or underuse it (sacrificing accuracy). By dynamically balancing accuracy preservation against computational cost, these results demonstrate that the policy achieves near-reference fidelity at a fraction of the cost of fully implicit integration.

\subsubsection{Comparison to Pure QSS Baseline}

While the QSS-only baseline offers 3--8$\times$ speedup over CVODE (Table~\ref{tab:0d_results}, CPU Time columns), it systematically fails to capture ignition accurately: QSS overpredicts ignition delay by 13--45\% relative to the CVODE reference. This unacceptable error stems from QSS's reliance on the quasi-steady-state approximation for radicals, which breaks down precisely during the rapid chain-branching phase that defines ignition. 

The RL-adaptive policy bridges this accuracy-efficiency gap by selectively invoking CVODE only where QSS fails. By learning to identify chemical regimes (via temperature, radical concentrations, and temporal gradients) rather than memorizing problem-specific patterns, the policy achieves CVODE-level accuracy at a cost closer to QSS than to CVODE. This represents a practical middle ground unattainable by either pure solver.

\subsection{One-Dimensional Counterflow Diffusion Flames}
\label{subsec:1d_results}

\subsubsection{Generalization of the 0D-Trained Policy}

\begin{figure*}[t]
    \centering
    \includegraphics[width=\textwidth]{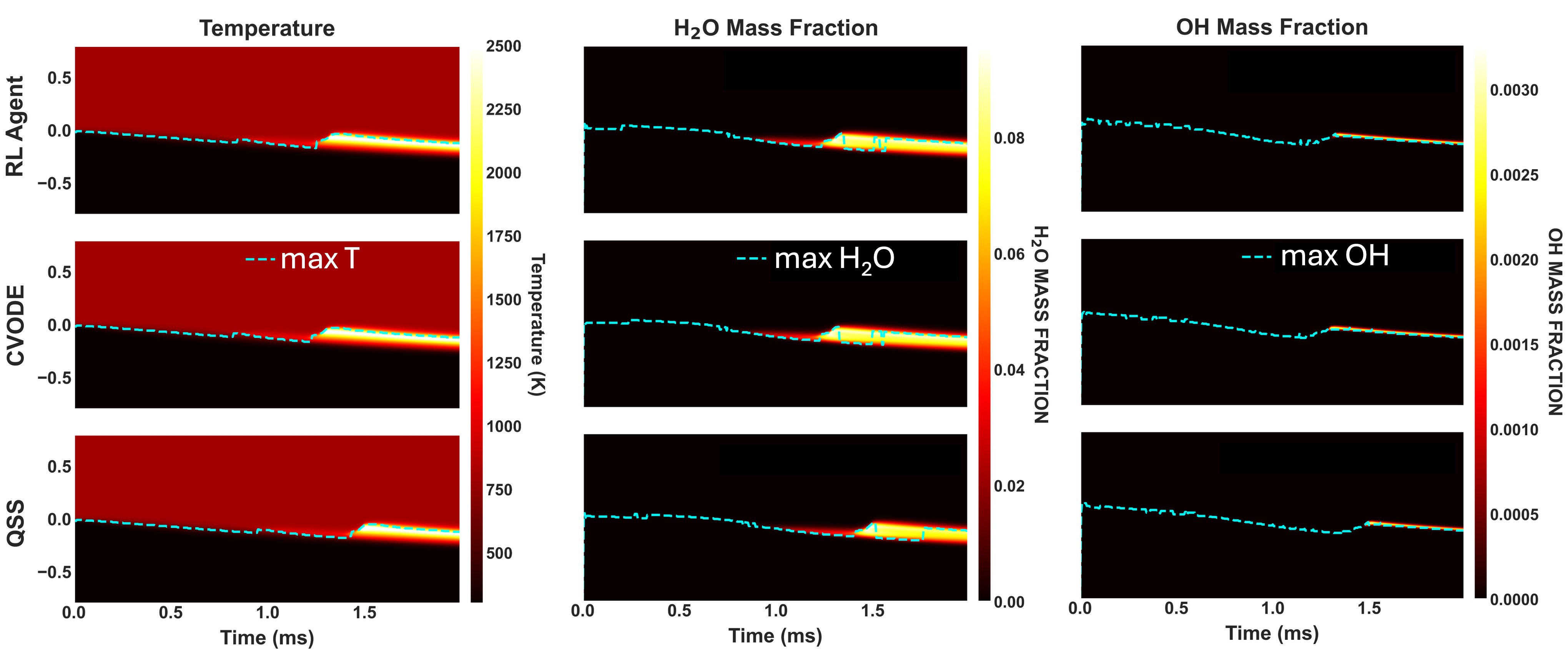}
    \caption{Space-time evolution of temperature, \ce{H2O}, and \ce{OH} illustrating agreement between the RL-adaptive solution and the CVODE reference, and deviations introduced by pure QSS.}
    \label{fig:flamefront}
\end{figure*}

To assess cross-domain generalization, we deployed the RL-based adaptive solver-selection policy trained \emph{exclusively} on 0D constant-pressure homogeneous reactors in a one-dimensional (1D) \textit{n}-dodecane/air counterflow diffusion flame, without any retraining or domain-specific tuning. These 1D simulations were carried out using the open-source code Ember \cite{LONG2018105}, which solves the coupled species, energy, and momentum equations utilizing the rebalanced Strang operator splitting method. The counterflow diffusion flame simulations employed a symmetric opposed-jet configuration, with \textit{n}-dodecane introduced on the fuel side at 300~K and air supplied on the oxidizer side at 800~K. Strain rates were varied between 10 and 2000~s$^{-1}$. The computational domain spanned $x\in[-8\times10^{-4},\,8\times10^{-4}]$~m and was initialized using a uniform axial grid of 500 points. Initial temperature and species profiles were constructed from a piecewise stoichiometric equilibrium profile smoothly matched to the prescribed inlet conditions.
This evaluation setting differs fundamentally from the training environment by introducing transport-chemistry coupling (advection and diffusion), steep spatial gradients, and strain-rate-dependent flame structure, all of which can induce data patterns not present in homogeneous ignition trajectories.

Despite these distribution shifts, the policy exhibits strong zero-shot transfer. Across all simulated strain rates ($10$--$2000~\mathrm{s}^{-1}$), it maintained numerical stability and preserved physical fidelity, achieving ignition-delay errors below $2.5\%$ and temperature root-mean-square (RMS) errors below $2.5~\mathrm{K}$ relative to a CVODE reference. These results suggest that the learned decision rule does not merely memorize homogeneous ignition behavior but instead exploits stiffness-relevant signatures encoded in the local thermochemical state and its gradient (as represented by the selected features) to invoke expensive implicit integration only when warranted.

\vspace{0.25cm}

\subsubsection{Spatiotemporal Analysis}

\begin{figure*}[htbp]
    \centering
    \includegraphics[width=\textwidth]{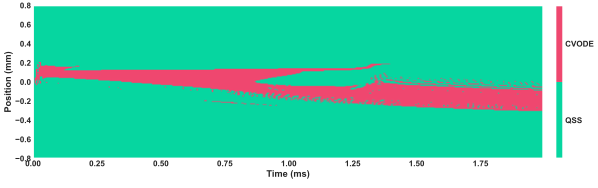}
    \caption{Spatiotemporal solver deployment in the 1D counterflow diffusion flame at strain rate $2000~\mathrm{s}^{-1}$, as selected by the RL policy. Red denotes CVODE and turquoise denotes QSS.}
    \label{fig:1d_spatiotemporal_SR_2000}
\end{figure*}

Although the underlying configuration is one-dimensional in space, Figures~\ref{fig:flamefront} and \ref{fig:1d_spatiotemporal_SR_2000} present space-time visualizations obtained by treating time as the second axis. This representation provides an interpretable view of both the flame-structure evolution and the associated solver allocation decisions. The results shown correspond to the representative high-strain case of $2000~\mathrm{s}^{-1}$.

Figure~\ref{fig:flamefront} depicts the space-time evolution (time increasing from right to left) of temperature (left), \ce{H2O} mass fraction (middle), and \ce{OH} mass fraction (right). The RL-adaptive solution closely reproduces the CVODE reference, capturing both the flame-front evolution and the spatial shift that occurs during ignition. In contrast, pure QSS exhibits delayed ignition, as seen by the delayed rise in \ce{OH} and a corresponding lag in temperature rise close to stoichiometric conditions, consistent with reduced accuracy during radical buildup in regions of high chemical reactivity.

The spatiotemporal solver map in Figure~\ref{fig:1d_spatiotemporal_SR_2000} illustrates how the RL policy allocates integration methods across the space-time domain. CVODE is selectively activated in a narrow band surrounding the flame front for the majority of the simulation, while QSS dominates elsewhere, indicating that the policy concentrates expensive implicit integration in regions of highly reactive chemistry and pronounced stiffness. Minor speckling is observed, which we attribute to stochasticity in training; however, the dominant structure is clear and physically consistent. Specifically, the policy localizes CVODE to regions of large thermal and species gradients and high chemical activity, precisely where numerical stiffness is expected to be strongest. Overall, CVODE is selected at only $\sim14\%$ of space-time points, thus implying the efficiency of the RL-derived solver allocation policy.

\vspace{0.25cm}
\subsubsection{Computational Performance and Efficiency}

\begin{table*}[htbp]
\centering
\caption{Performance metrics for RL-adaptive solver selection in 1D counterflow diffusion flames compared to CVODE and QSS baselines. All cases use n-dodecane mechanism with 106 species.}
\label{tab:1d_metrics}
\begin{adjustbox}{width=\textwidth}
\begin{tabular}{c|ccc|ccc|cccc}
\hline
& \multicolumn{3}{c|}{\textbf{Ignition Delay}} 
& \multicolumn{3}{c|}{\textbf{CPU Time}} 
& \multicolumn{4}{c}{\textbf{Performance \& Accuracy}} \\
\cline{2-11}
\textbf{Strain Rate} 
& \textbf{CVODE} & \textbf{QSS} & \textbf{RL} 
& \textbf{CVODE} & \textbf{QSS} & \textbf{RL} 
& \textbf{Speedup} & \textbf{QSS} & \textbf{RL} & \textbf{CVODE} \\
\textbf{(s$^{-1}$)}
& \textbf{(ms)} & \textbf{(ms)} & \textbf{(ms)} 
& \textbf{(s)} & \textbf{(s)} & \textbf{(s)} 
& \textbf{vs CVODE} & \textbf{RMSE (K)} & \textbf{RMSE (K)} & \textbf{Usage (\%)} \\
\hline
10 & 1.47 & 1.63 & 1.49 & 78.32 & 8.32 & 33.98 & 2.30$\times$ & 58.93 & 10.49 & 15.4 \\
100 & 1.41 & 1.58 & 1.43 & 79.45 & 8.57 & 35.54 & 2.24$\times$ & 90.22 & 4.48 & 15.6 \\
500 & 1.27 & 1.43 & 1.30 & 79.33 & 8.64 & 35.66 & 2.22$\times$ & 101.73 & 1.09 & 15.4 \\
1000 & 1.20 & 1.36 & 1.23 & 86.32 & 8.95 & 37.96 & 2.27$\times$ & 14.33 & 2.21 & 14.3 \\
2000 & 1.21 & 1.41 & 1.24 & 80.78 & 9.29 & 35.01 & 2.31$\times$ & 12.93 & 1.34 & 12.1 \\
\hline
\end{tabular}
\end{adjustbox}
\end{table*}

Table~\ref{tab:1d_metrics} reports runtime and accuracy metrics across strain rates ranging from $10$ to $2000~\mathrm{s}^{-1}$. Overall, the RL-adaptive strategy delivers a consistent speedup of approximately $2.2\times$ relative to CVODE while maintaining numerical stability and near-reference accuracy. For example, at $1000~\mathrm{s}^{-1}$ the total CPU time is reduced from $86.3~\mathrm{s}$ (CVODE) to $38.0~\mathrm{s}$ (RL-adaptive). The inference overhead associated with evaluating the policy network and issuing solver-selection decisions remains negligible, contributing only $2.7\%$ of the total wall-clock time.

Across strain rates, solver utilization patterns remain stable: the implicit CVODE integrator is selected at only $12$--$15\%$ of space-time points, with the remaining points advanced using the cheaper QSS update. While the standalone QSS solver achieves the lowest runtime, it does so at the expense of substantially degraded accuracy compared to the RL-adaptive method. It should also be noted that the accuracy-cost trade-off of the RL policy is directly controlled through the training constraint (or tolerance) used to define acceptable error, as described in section \ref{sec:reward}. Looser tolerances encourage more frequent selection of QSS, whereas tighter tolerances increase reliance on CVODE (perhaps even beyond very stiff regions). In this sense, the proposed approach provides a principled continuum between the two extremes (always-CVODE versus always-QSS). Depending on user preference, the obtained policy can be tuned to provide higher CVODE-like accuracy (at higher computational cost), or to approach QSS speeds (with higher ignition delay errors). Therefore, the proposed framework offers a pathway to construct policies that can be tuned to lie on different points along the accuracy-efficiency spectrum.

\section{Conclusion}
\label{sec:conclusion}

This study introduced a reinforcement learning (RL) framework for adaptive solver selection in detailed combustion chemistry, featuring automatic switching between an implicit integrator (CVODE) and a quasi-steady-state (QSS) method. By casting solver selection as a Markov decision process and training under a Lagrangian, constraint-aware objective, the policy learns to reduce computational cost while enforcing stringent accuracy requirements. When evaluated on zero-dimensional (0D) homogeneous reactor trajectories, the learned policy achieves speedups ranging from $1.11\times$ to $10.58\times$ relative to CVODE, while maintaining ignition-delay errors below $2.6\%$. Although trained solely on zero-dimensional (0D) homogeneous reactor data, the learned policy transfers effectively to one-dimensional (1D) \textit{n}-dodecane/air counterflow diffusion flames, achieving $\approx 2.2\times$ speedup across strain rates while preserving near-reference accuracy (ignition-delay errors below $2.5\%$). The observed transferability indicates that the policy captures stiffness-relevant signatures embedded in local thermochemical states and gradients (in particular temperature, radical species levels, and temporal-gradient proxies) rather than memorizing trajectory-specific patterns.

Spatiotemporal solver-allocation maps further reveal physically interpretable behavior: the policy concentrates implicit integration in narrow high-stiffness regions near the flame front and relies on QSS elsewhere. This adaptive partitioning emerges without hand-crafted switching rules, thereby demonstrating the potential of data-driven control in regimes where stiffness, multiscale kinetics, and evolving thermochemical conditions limit the effectiveness of fixed heuristics. Moreover, policy evaluation incurs minimal overhead (less than $3\%$ of total runtime), making the approach compatible with existing operator-splitting workflows in production CFD solvers.

Several limitations remain, providing a pathway for future extensions. The present policy was trained using a single fuel mechanism (\textit{n}-dodecane) and would be extended to multi-fuel formulations in a future study. Scaling to large 2D and 3D turbulent configurations will also require efficient batched inference and load balancing, and the lightweight network architecture used in this study makes these practical. Finally, the current formulation focuses on binary solver selection; future work will expand the action space to include solvers with variable tolerances and reduced-order chemistry options to enable finer control of the accuracy-efficiency trade-off.

By allowing learned policies to govern solver adaptivity, this work points toward simulation frameworks that satisfy desired accuracy constraints while learning to optimize their computational strategies for efficiency.

\bibliographystyle{unsrtnat}
\bibliography{references} 

\end{document}